\documentclass[11pt]{article}

\usepackage[utf8]{inputenc}
\usepackage{lmodern}
\usepackage[T1]{fontenc}
\usepackage{amsmath,amssymb}
\usepackage{graphicx}
\usepackage{booktabs}
\usepackage{xcolor}
\usepackage{tikz}
\usetikzlibrary{positioning, arrows.meta, fit, backgrounds}
\usepackage{hyperref}
\usepackage[numbers,sort&compress]{natbib}
\usepackage{geometry}
\hypersetup{colorlinks=true, linkcolor=blue, citecolor=blue, urlcolor=blue}

\title{\textbf{Quantum Circuits in Diffusion Models:\\
A Fair-Comparison Study and a Mechanistic\\
Analysis of Angle-Embedding Failures}}

\author{
  Jaeuk Kim \qquad Sanghoon Yoo\\[2pt]
  \small NextITS Co., Ltd.\\
  \small \texttt{freak91uk@hnextits.com, stmlshu@hnextits.com}
}
\date{}

\begin{document}
\maketitle

\begin{abstract}
We study the integration of variational quantum circuits (VQCs) into diffusion
models through a squeeze-and-excitation (SE) channel-modulation scaffold that
isolates the quantum contribution. Using a \emph{role-matched} classical control
(a classical MLP that plays the identical role in the same scaffold) and
multi-seed significance testing across DDPM and latent diffusion on MNIST and
CIFAR-10 (with a score-based NCSN study on MNIST), both quantum cores achieve
comparable mean FID to the role-matched classical control across DDPM and latent
diffusion, while paired sampling-seed tests for EfficientSU2 detect no
statistically significant difference. While the
quantum cores use $4.5$--$9\times$ fewer parameters than this control, at matched
budgets they attain slightly lower mean FID in all four MNIST/CIFAR-10 comparisons;
the differences are small and not significance-tested, so the experiments do not
establish a \emph{quantum parameter-efficiency advantage}. We further mechanistically identify and resolve
a structural failure in the score-based NCSN setting:
the unbounded score target ($\propto 1/\sigma$) drives angle-embedding inputs far
beyond the $2\pi$ period of the rotation gates, a phase-\emph{aliasing} effect that
flattens the expectation-value landscape and collapses the quantum modulator.
A bounding transformation $\theta \leftarrow \pi\tanh(\cdot)$ projects the feature
space onto the non-aliasing domain $(-\pi,\pi)$, removing the failure and
substantially improving both quantum cores; normalized
RealAmplitudes attains the lowest mean FID in this single-training-run
comparison (not established as statistically significant). Because all circuits are classically simulated
at a few-qubit scale, we characterize whether a variational quantum
parameterization is a useful inductive bias under strictly matched controls,
rather than claiming quantum advantage; under this controlled comparison the
quantum core exhibits \emph{functional parity}---not superiority---with a
classical core of equal parameter budget. Our contributions are (i)
a rigorous fair-comparison methodology for quantum-enhanced generative models and
(ii) a mechanistic account of when and why angle embeddings fail.
\end{abstract}

\section{Introduction}
\label{sec:intro}
Quantum machine learning (QML)~\citep{cerezo2021vqa} for generative modeling has
attracted strong interest, with a growing body of work inserting parameterized
quantum circuits into autoencoders, GANs, and, more recently, diffusion
models~\citep{zhang2024quddpm,parigi2024qndgdm}.
A recurring difficulty in this literature is
evaluation: a quantum-augmented model is often compared against a plain baseline
without a classical module of matched capacity in the same location, so an
observed gain may simply reflect the extra parameters and nonlinearity of the
inserted block rather than anything quantum. When the quantum circuit is small
enough to be classically simulable, as is the case at the few-qubit scale used
in practice, the experiments are not designed to test computational quantum
advantage; what remains to be measured is whether the quantum
parameterization is a \emph{useful inductive bias} relative to a classical core
playing the same role.

We therefore adopt a deliberately conservative experimental design. We fix a
single squeeze-and-excitation (SE) modulation scaffold and vary only its
\emph{core}, swapping a variational quantum circuit for a \emph{role-matched}
classical multilayer perceptron (MLP)---one that occupies the identical position
in the same scaffold, though with \emph{more} core parameters (144 vs.\ 16/32),
making it a \emph{higher-parameter}, not smaller, control---or for nothing at all, with
all wrappers, initializations, and training schedules held identical. Differences are then
assessed with multi-seed significance testing rather than single-run point
estimates. Diffusion models are an attractive testbed because the same scaffold
can be evaluated across three structurally distinct families --- denoising
diffusion (DDPM), latent diffusion (LDM), and score-based generation (NCSN) ---
on standard image benchmarks.

Across this matrix we find no statistically detectable difference: quantum cores
match the higher-parameter classical control while using $4.5$--$9\times$ fewer parameters
\emph{within the core}, with no significant win or loss on DDPM or LDM. We stress
that this efficiency is local to the modulation core, which is a tiny fraction of
the full $\sim$23\,M-parameter model (Sec.~\ref{sec:exp}); it is a statement
about a compact parameterization, not about whole-model size, runtime, or memory.
The one exception, score-based NCSN, turns out to be informative: we trace the
dominant observed failure to aliasing in the periodic angle embedding when fed
the unbounded score target---rather than to an obvious shortage of core
parameters---a failure we diagnose and fix. We
make no claim of quantum advantage; the value of the work is a fair-comparison
methodology and a transferable mechanistic insight.

\paragraph{Contributions.}
\begin{itemize}
  \item A \textbf{fair-comparison framework} (Sec.~\ref{sec:method}): a fixed SE
  scaffold in which only the \emph{core} is swapped between a quantum circuit, a
  role-matched classical MLP (with \emph{more} core parameters), or nothing, with
  identity initialization so all variants share the same starting point.
  \item A \textbf{no-difference finding with no efficiency advantage}
  (Sec.~\ref{sec:exp}): across DDPM and latent diffusion on MNIST/CIFAR-10,
  quantum cores match the higher-parameter ($144$-parameter) classical control
  ($|\mathrm{SNR}|<0.3$); but a \emph{parameter-matched} classical core
  ($16$/$32$ parameters) is numerically comparable to the quantum core, so the
  apparent ``fewer-parameter'' edge is not a significant quantum efficiency advantage.
  \item A \textbf{mechanistic discovery} (Sec.~\ref{sec:aliasing}): score-based
  (NCSN) models fail with quantum cores due to angle-embedding aliasing; a
  $\pi\tanh$ normalization substantially improves both quantum cores and, for
  RealAmplitudes, attains the lowest mean FID in a single-training-run comparison.
  \item \textbf{Methodological rigor over empirical confounding}
  (Sec.~\ref{sec:method}): rather than reporting unvetted gains, we decouple
  architectural confounders from quantum expressivity, establishing a
  transparent, reproducible evaluation standard for hybrid generative models and
  isolating a transferable mechanistic failure mode.
\end{itemize}

\section{Background}
\label{sec:bg}
\subsection{Diffusion models}
A diffusion model defines a forward process that gradually adds Gaussian noise,
$x_t=\sqrt{\bar\alpha_t}\,x_0+\sqrt{1-\bar\alpha_t}\,\epsilon$ with
$\epsilon\sim\mathcal N(0,I)$, and learns a reverse (denoising) process that
predicts $\epsilon$~\citep{ho2020ddpm}. Score-based models instead learn the
score $\nabla_x\log p(x)$ over a range of noise scales~\citep{song2019ncsn,
song2021sde}. Latent diffusion runs the process in a compressed latent
space~\citep{rombach2022ldm}.

\subsection{Variational quantum circuits and angle embedding}
A VQC encodes classical inputs as rotation angles (angle embedding), applies
parameterized entangling layers, and reads out expectation values such as
$\langle Z_i\rangle$~\citep{schuld2020circuit,bergholm2018pennylane}. Crucially,
rotation gates are $2\pi$-periodic, a fact central to Sec.~\ref{sec:aliasing}.

\section{Method: Quantum SE Modulation with Fair Controls}
\label{sec:method}

\subsection{The SE scaffold}
We insert a squeeze-and-excitation (SE) channel gate~\citep{hu2018senet} at the
U-Net bottleneck. Given features $x\in\mathbb{R}^{B\times C\times H\times W}$, we
pool over space to $v\in\mathbb{R}^{B\times C}$, project to $n_q$ values, apply a
\emph{core} $f_\theta:\mathbb{R}^{n_q}\!\to\!\mathbb{R}^{n_q}$, map back to a
per-channel gate, and modulate:
\begin{equation}
  g = 1 + \tanh\!\big(W_{\uparrow}\,\tanh(f_\theta(W_{\downarrow}\,v))\big),
  \qquad x' = x \odot g,
\end{equation}
where $W_{\downarrow}$ and $W_{\uparrow}$ are the down/up projections. The
up-projection $W_{\uparrow}$ is zero-initialized, so $g\equiv 1$ and $x'=x$ at
initialization; every variant therefore starts from the identical, unmodulated
network and the gate is wrapped residually. The \emph{only} object that changes
between conditions is the core $f_\theta$.

\begin{figure}[t]
  \centering
  \begin{tikzpicture}[
      font=\small,
      box/.style={draw, rounded corners=2pt, minimum height=8mm, inner sep=3pt, align=center},
      core/.style={draw, rounded corners=2pt, minimum height=8mm, minimum width=20mm, align=center, fill=yellow!18, very thick},
      op/.style={draw, circle, inner sep=1pt, minimum size=5mm},
      >={Stealth[length=2mm]},
      node distance=6mm and 6mm,
    ]
    \node[box] (x) {$x$\\\scriptsize $C{\times}H{\times}W$};
    \node[box, right=of x] (pool) {GAP\\\scriptsize $\to v$};
    \node[box, right=of pool] (down) {$W_{\downarrow}$\\\scriptsize $C\!\to\! n_q$};
    \node[core, right=of down] (core) {\textbf{core} $f_\theta$};
    \node[box, right=of core] (up) {$W_{\uparrow}$\\\scriptsize 0-init};
    \node[box, right=of up] (gate) {$1{+}\tanh$};
    \node[op, right=of gate] (mul) {$\odot$};
    \node[box, right=of mul] (out) {$x'$};

    \draw[->] (x) -- (pool);
    \draw[->] (pool) -- (down);
    \draw[->] (down) -- (core);
    \draw[->] (core) -- (up);
    \draw[->] (up) -- (gate);
    \draw[->] (gate) -- (mul);
    \draw[->] (mul) -- (out);
    \draw[->] (x.south) to[out=-90, in=-90] (mul.south);

    \node[below=10mm of core, align=center, font=\scriptsize] (variants)
      {\texttt{se\_mode}: \textbf{quantum} (VQC, 16/32 p.) $\;|\;$
       \textbf{classical} (MLP, 144 p.) $\;|\;$ \textbf{none}};
    \draw[densely dashed, ->] (variants) -- (core);
  \end{tikzpicture}
  \caption{Fair-comparison scaffold at the U-Net bottleneck. Pooled channel
  statistics $v$ are projected to $n_q$ values, transformed by a \emph{core}
  $f_\theta$, projected back through a zero-initialized $W_\uparrow$ (so the gate
  is the identity at init), and used to modulate $x$ residually. The
  \texttt{se\_mode} axis swaps \emph{only} the core --- a variational quantum
  circuit, a role-matched classical MLP, or nothing --- holding the wrapper,
  initialization, and training schedule fixed.}
  \label{fig:arch}
\end{figure}
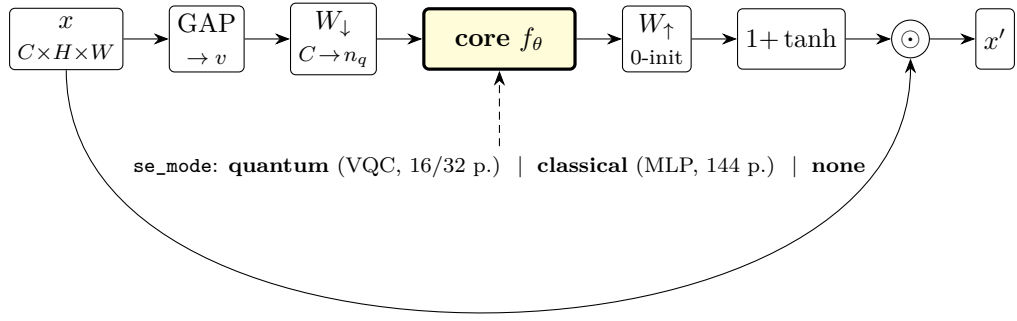

\subsection{The \texttt{se\_mode} axis and the role-matched control}
The core is one of:
\begin{itemize}
  \item \texttt{none}: no SE module (plain diffusion baseline).
  \item \texttt{classical}: an MLP core in the \emph{same role} (144 core
  parameters) --- the control. It has \emph{more} core parameters than either
  quantum core, so it is a \emph{higher-parameter}, not smaller, comparison point.
  \item \texttt{quantum}: a VQC core, either RealAmplitudes (16 parameters) or
  EfficientSU2 (32 parameters).
\end{itemize}
The quantum cores thus use $4.5$--$9\times$ \emph{fewer} core parameters than the
classical control; matching a higher-parameter control is the sense in which
the quantum parameterization is compact. We emphasize that this comparison is
\emph{role-matched}, not parameter-matched: a parameter-matched comparison would
shrink the classical core to $16/32$ parameters, which we examine separately in
Sec.~\ref{sec:exp} (tiny-MLP control).

\subsection{Evaluation protocol}
For each configuration we generate samples under $5$ generation seeds and report
FID~\citep{heusel2017fid} and SSIM~\citep{wang2004ssim} as mean$\pm$std. We summarize each paired comparison with an
\emph{effect size} $\mathrm{SNR}=\overline{\Delta}/\mathrm{std}(\Delta)$ over
seeds---a standardized mean difference (Cohen's-$d$ family), not a $p$-value. It
is the per-seed analogue of a paired $t$ statistic, $t=\mathrm{SNR}\sqrt{n}$ with
$n{=}5$ ($\mathrm{df}{=}4$), so our flag $|\mathrm{SNR}|>2$ corresponds to
$|t|>4.5$ (two-sided $p\!\approx\!0.01$) and is a conservative threshold. These seeds vary
the \emph{generation} (sampling) noise at a fixed checkpoint; they probe ranking
stability under sampling noise. To also probe training-initialization variance, we
retrain the two SE cores under four additional seeds, for a total of five
independent training seeds, and re-evaluate each checkpoint under five generation
seeds (Table~\ref{tab:trainseed}). SSIM is computed
between \emph{arbitrarily paired} generated and real images (no correspondence
exists for unconditional generation); we therefore treat it only as a coarse
structural-similarity indicator and rely on FID for quality.

\section{Experiments}
\label{sec:exp}

\paragraph{Setup.} All cores use $8$-qubit circuits (PennyLane
\texttt{default.qubit}, backprop) and are trained for $300$ epochs with Adam
(learning rate $10^{-4}$, batch size $64$), a cosine-annealed schedule, gradient
clipping at $1.0$, and an exponential moving average of weights (decay $0.9999$)
used for evaluation. Hyperparameters are identical across cores; only the
\texttt{se\_mode} differs. We evaluate on MNIST and CIFAR-10, computing FID with
\texttt{pytorch-fid} and SSIM with \texttt{torchmetrics} over $1000$ generated
and $1000$ real images ($500$ for NCSN), using $100$ sampling steps and $5$
generation seeds. Full hyperparameters and SSIM tables are in the Appendix.

\begin{table}[t]
  \centering
  \caption{Parameter accounting (DDPM/MNIST). The \texttt{se\_mode} core is the
  only object that changes; it is a tiny fraction of the model. The quantum vs.\
  classical \emph{core} difference is $\sim$$10^2$ parameters out of $\sim$23\,M
  ($\approx0.001\%$), so ``$4.5$--$9\times$ fewer core parameters'' is a statement about
  a compact component, not the whole model.}
  \label{tab:params}
  \small
  \begin{tabular}{lccc}
    \toprule
    \texttt{se\_mode} & core params (per block) & SE-module params & total trainable\\
    \midrule
    none         & 0   & 0       & $23{,}328{,}641$\\
    classical-SE & 144 & $134{,}704$ & $23{,}463{,}345$\\
    quantum-real & 16  & $134{,}448$ & $23{,}463{,}089$\\
    quantum-su2  & 32  & $134{,}480$ & $23{,}463{,}121$\\
    \bottomrule
  \end{tabular}
\end{table}

\begin{table}[t]
  \centering
  \caption{Multi-seed FID (mean$\pm$std, 5 seeds; lower is better). Quantum
  cores perform comparably to the higher-parameter (role-matched) classical SE
  control on DDPM and latent diffusion. NCSN is the exception, analyzed in
  Sec.~\ref{sec:aliasing}.}
  \label{tab:main}
  \small
  \begin{tabular}{llcccc}
    \toprule
    Model & Dataset & none & classical-SE & quantum-real & quantum-su2\\
    & & (0 p.) & (144 p.) & (16 p.) & (32 p.)\\
    \midrule
    DDPM & MNIST   & $14.79\pm0.77$ & $14.59\pm0.86$ & $14.80\pm0.74$ & $\mathbf{14.52\pm0.70}$\\
    DDPM & CIFAR-10& $55.14\pm0.57$ & $54.85\pm0.45$ & $54.96\pm0.92$ & $\mathbf{54.74\pm0.60}$\\
    LDM  & MNIST   & $15.37\pm0.51$ & $14.88\pm0.32$ & $15.33\pm0.54$ & $\mathbf{14.85\pm0.33}$\\
    LDM  & CIFAR-10& $99.91\pm0.82$ & $99.45\pm0.88$ & $\mathbf{98.94\pm0.65}$ & $99.18\pm0.25$\\
    NCSN & MNIST   & $23.73\pm0.60$ & $\mathbf{19.07\pm0.66}$ & $24.52\pm0.45$ & $24.54\pm0.56$\\
    \bottomrule
  \end{tabular}
\end{table}

\begin{table}[t]
  \centering
  \caption{Paired effect size, quantum-su2 vs.\ classical-SE. We define
  $\Delta=\mathrm{FID}_{\text{classical}}-\mathrm{FID}_{\text{quantum}}$ per seed
  (positive favors the quantum core), and \emph{wins} counts seeds on which the
  quantum core has the lower FID;
  $\mathrm{SNR}=\overline{\Delta}/\mathrm{std}(\Delta)$ over 5 seeds (equivalently
  a paired $t=\mathrm{SNR}\sqrt5$, $\mathrm{df}{=}4$).
  $|\mathrm{SNR}|<2$ ($|t|<4.5$, $p\!\gtrsim\!0.01$) everywhere on DDPM/LDM
  $\Rightarrow$ no detectable difference.}
  \label{tab:snr}
  \small
  \begin{tabular}{llccc}
    \toprule
    Model & Dataset & wins & mean gap (FID) & SNR\\
    \midrule
    DDPM & MNIST    & 3/5 & $+0.072\pm0.434$ & $0.17$\\
    DDPM & CIFAR-10 & 3/5 & $+0.116\pm0.629$ & $0.18$\\
    LDM  & MNIST    & 3/5 & $+0.027\pm0.498$ & $0.05$\\
    LDM  & CIFAR-10 & 3/5 & $+0.271\pm0.961$ & $0.28$\\
    NCSN & MNIST    & 0/5 & $-5.471\pm1.040$ & $\mathbf{-5.26}$\\
    \bottomrule
  \end{tabular}
\end{table}

\paragraph{Result: comparable to the higher-parameter control.}
On DDPM and latent diffusion (both datasets) quantum cores are statistically
indistinguishable from the classical-SE control (Table~\ref{tab:snr},
$|\mathrm{SNR}|<0.3$; the paired test targets quantum-su2, and quantum-real shows
comparable mean FID in Table~\ref{tab:main}), while using $4.5$--$9\times$ fewer core parameters. As we
show next, however, this does not establish a quantum efficiency
advantage: parameter-matched classical controls achieve similar mean FID at the
same $16$/$32$-parameter budgets. NCSN is
the lone significant case ($\mathrm{SNR}=-5.26$, quantum \emph{worse});
we diagnose it in Sec.~\ref{sec:aliasing}.

\begin{figure}[t]
  \centering
  \includegraphics[width=0.92\linewidth]{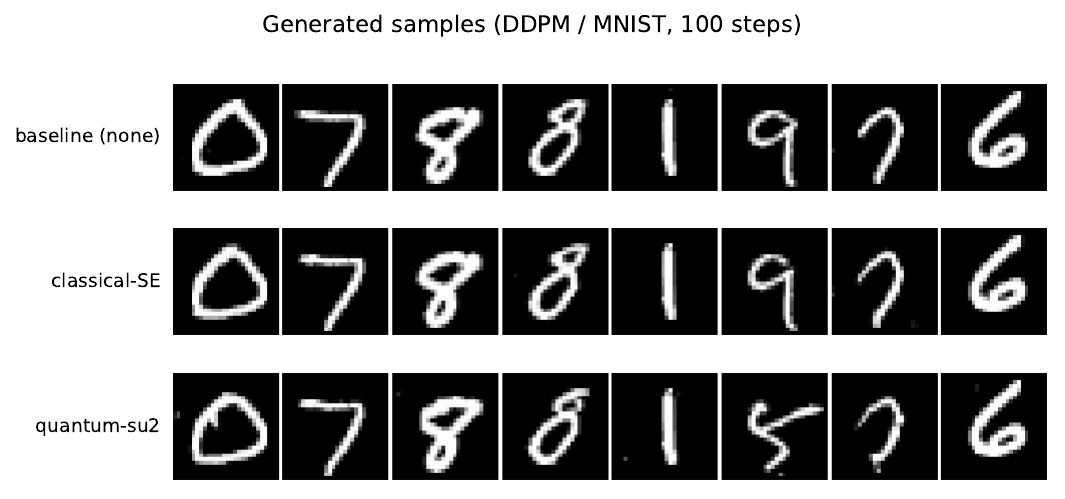}
  \caption{Generated samples (DDPM/MNIST, $100$ steps, shared seed). Sample
  quality is visually comparable across the no-SE baseline, the classical-SE
  control, and the quantum-su2 core, consistent with the comparable FID results in
  Table~\ref{tab:main}.}
  \label{fig:samples}
\end{figure}

\paragraph{Is the efficiency real? A parameter-matched control.}
The headline comparison above pits a $16$/$32$-parameter quantum core against a
$144$-parameter classical-SE core, so ``fewer parameters'' conflates
\emph{quantumness} with \emph{smallness}. To separate them we add classical cores
shrunk to the \emph{same} budget via a low-rank bottleneck
($16$ and $32$ parameters; Table~\ref{tab:tinymlp}). At matched budget the two
cores are within sampling noise of each other: the quantum core is marginally
lower at $16$ parameters ($14.80$ vs.\ $15.18$) and at $32$ ($14.52$ vs.\
$14.67$), gaps comparable to the per-seed standard deviation ($\sim$$0.7$--$0.9$).
Moreover the $144$-parameter classical-SE ($14.59$) is no better than its
$32$-parameter version ($14.67$), so the $144$-parameter control was simply
over-provisioned for this task. We therefore find \emph{no significant
parameter-efficiency advantage either way}: at equal budget the quantum and
classical cores obtain numerically comparable mean FID (the quantum core is not worse, and
marginally lower here), and on DDPM/MNIST the SE core barely moves FID from the
no-SE baseline ($14.79$) regardless of its form.

\begin{table}[t]
  \centering
  \caption{Parameter-matched control (DDPM/MNIST, FID, 5 seeds). At equal core
  budget the quantum and classical cores are within sampling noise (the quantum
  core marginally lower at both $16$ and $32$); the $144$-parameter classical-SE
  is no better than the $32$-parameter one. No statistically significant
  efficiency advantage is established.}
  \label{tab:tinymlp}
  \small
  \begin{tabular}{lcc}
    \toprule
    Core & params & FID\\
    \midrule
    none (baseline)   & 0   & $14.79\pm0.77$\\
    quantum-real      & 16  & $\mathbf{14.80\pm0.74}$\\
    classical (rank-1)& 16  & $15.18\pm0.89$\\
    quantum-su2       & 32  & $\mathbf{14.52\pm0.70}$\\
    classical (rank-2)& 32  & $14.67\pm0.72$\\
    classical-SE      & 144 & $14.59\pm0.86$\\
    \bottomrule
  \end{tabular}
\end{table}

\paragraph{Generalization to CIFAR-10.} The same parameter-matched comparison on
DDPM/CIFAR-10 (Table~\ref{tab:tinymlpcifar}) gives a similar numerical pattern:
the quantum core is marginally lower at $16$ parameters ($54.96$ vs.\ $55.30$)
and at $32$ parameters ($54.74$ vs.\ $54.80$). Across MNIST and CIFAR-10, the
quantum cores obtain slightly lower mean FID in all four matched-budget
comparisons, but the differences are small and are not established as
statistically significant.

\begin{table}[t]
  \centering
  \caption{Parameter-matched control on DDPM/CIFAR-10 (FID, 5 seeds). As on
  MNIST, the quantum cores attain slightly lower mean FID at both $16$- and
  $32$-parameter budgets; however, the gaps are small relative to sampling
  variability and are not significance-tested.}
  \label{tab:tinymlpcifar}
  \small
  \begin{tabular}{lcc}
    \toprule
    Core & params & FID\\
    \midrule
    none (baseline)    & 0   & $55.14\pm0.57$\\
    quantum-real       & 16  & $\mathbf{54.96\pm0.92}$\\
    classical (rank-1) & 16  & $55.30\pm0.45$\\
    quantum-su2        & 32  & $\mathbf{54.74\pm0.60}$\\
    classical (rank-2) & 32  & $54.80\pm0.23$\\
    classical-SE       & 144 & $54.85\pm0.45$\\
    \bottomrule
  \end{tabular}
\end{table}

\paragraph{Robustness to training initialization.} The significance tests above
vary only the generation seed. To check that this comparability is not an artifact of one
lucky initialization, we retrain classical-SE and quantum-su2 under four
additional training seeds (five in total) and re-evaluate (each entry still
averaged over $5$ generation seeds; Table~\ref{tab:trainseed}). The two cores
track each other across all five training seeds (classical-SE spans
$14.59$--$15.04$, quantum-su2 $14.41$--$15.35$), with overlapping spread and no
consistent winner. The comparability is thus stable to training initialization, not
merely to sampling noise.

\begin{table}[t]
  \centering
  \caption{Training-seed robustness (DDPM/MNIST, FID; each cell is mean$\pm$std
  over 5 generation seeds). Across five independent training seeds the
  quantum and classical cores show overlapping performance ranges with no consistent winner.}
  \label{tab:trainseed}
  \small
  \begin{tabular}{lccccc}
    \toprule
    Core & seed 0 & seed 1 & seed 2 & seed 3 & seed 4\\
    \midrule
    classical-SE & $14.59\pm0.86$ & $15.04\pm0.93$ & $14.65\pm0.83$ & $14.70\pm0.82$ & $14.63\pm0.75$\\
    quantum-su2  & $14.52\pm0.70$ & $15.03\pm0.87$ & $14.41\pm0.70$ & $15.35\pm0.77$ & $14.46\pm0.64$\\
    \bottomrule
  \end{tabular}
\end{table}

\section{The Angle-Aliasing Failure in Score-Based Models}
\label{sec:aliasing}

\paragraph{Symptom.} On NCSN the quantum core is significantly worse than the
classical control (Table~\ref{tab:snr}), unlike every other configuration.

\begin{figure}[t]
  \centering
  \includegraphics[width=0.82\linewidth]{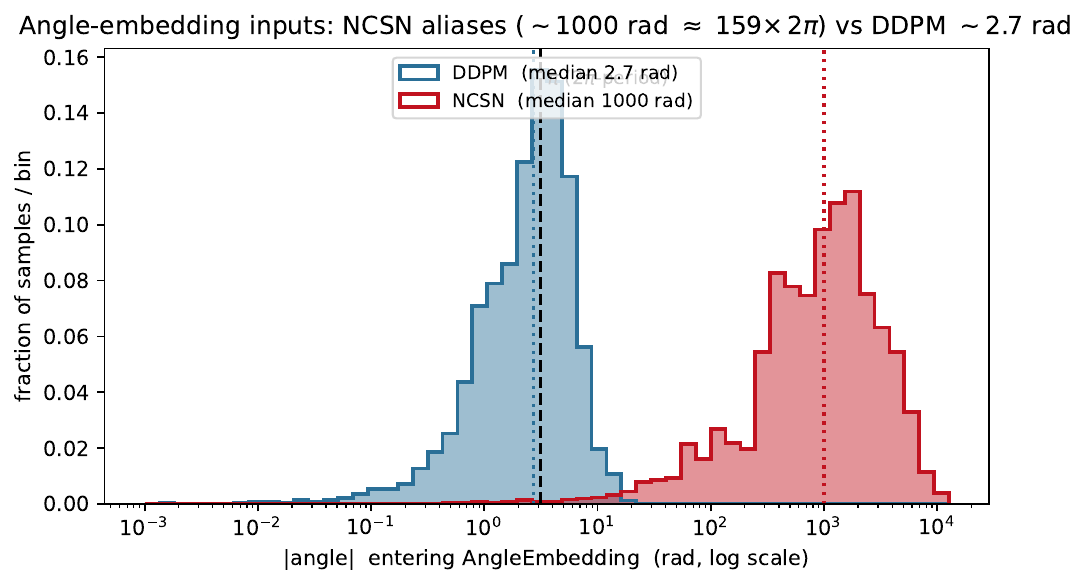}
  \caption{Magnitudes of the angles entering \texttt{AngleEmbedding} in a trained
  quantum-SE model (fraction of samples per log-spaced bin). DDPM activations
  remain $\mathcal{O}(1)$ (median $2.7$~rad), whereas NCSN's unbounded $1/\sigma$
  target pushes the inputs to a median of $\sim\!10^3$~rad
  ($\approx 159\times 2\pi$), deep into the aliasing regime. The dashed line
  marks $\pi$.}
  \label{fig:angles}
\end{figure}

\begin{figure}[t]
  \centering
  \includegraphics[width=0.78\linewidth]{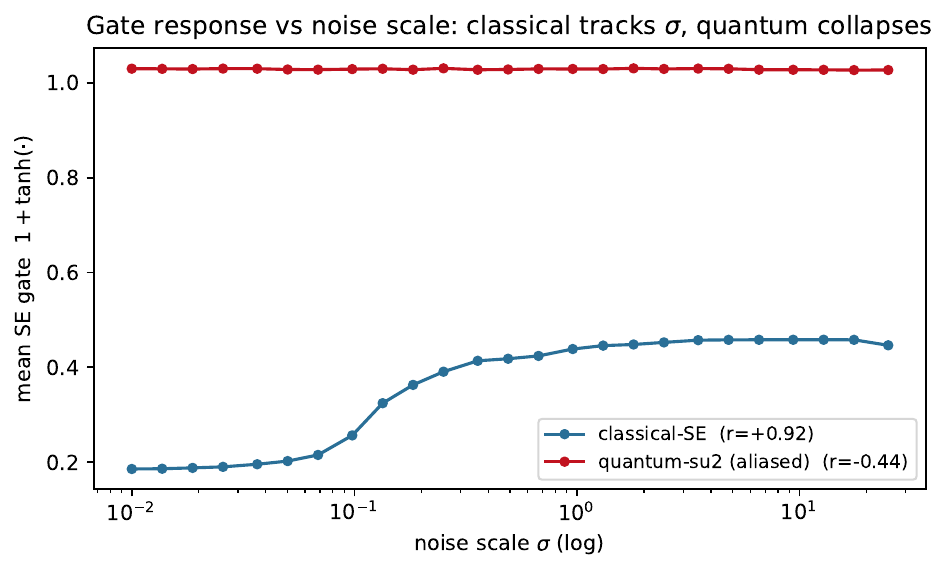}
  \caption{Consequence of aliasing: the batch- and channel-averaged SE gate
  $1+\tanh(\cdot)$, measured on held-out samples of the trained model, vs.\ NCSN
  noise scale $\sigma$. The classical core learns a monotone per-noise-scale
  schedule (correlation $r{=}{+}0.92$, gate range $0.27$); the aliased quantum
  core's gate spans only $0.004$ across the entire $\sigma$ range ($75\times$
  smaller), so although its correlation is nonzero ($r{=}{-}0.44$) the
  \emph{absolute} variation has collapsed and it cannot modulate by noise level.}
  \label{fig:gate}
\end{figure}

\paragraph{Diagnosis.} NCSN regresses an unbounded score target
($\propto 1/\sigma$). Measuring the actual arguments entering the angle
embedding of a trained EfficientSU2 model (Fig.~\ref{fig:angles}), the NCSN
inputs have a median magnitude of $\sim\!10^3$ radians (median $1000$~rad
$\approx 159\times 2\pi$; $99.6\%$ exceed $\pi$, with a maximum of
$\sim\!1.3\times10^4$~rad), i.e.\ they wrap around the $2\pi$ gate period
hundreds of times. By contrast the bounded DDPM activations stay
$\mathcal{O}(1)$ (median $2.7$~rad, at most a few wraps). The two distributions
are separated by roughly $2.5$ orders of magnitude with negligible overlap. At
the NCSN scale the periodic encoder \emph{aliases}, and the consequence is
directly visible in the gate it drives (Fig.~\ref{fig:gate}): the classical SE
gate varies smoothly and monotonically with the noise scale $\sigma$
(correlation $r{=}{+}0.92$, range $0.27$), having learned a per-noise-scale
schedule, whereas the quantum gate is essentially \emph{flat} (range $0.004$,
$\sim$$75\times$ smaller; $r{=}{-}0.44$)---it has collapsed and no longer tracks
$\sigma$. The failure is thus a property of the encoder's periodicity meeting an
unbounded target, not of the quantum core's capacity.

\paragraph{Fix.} We bound the embedding arguments with
$a \leftarrow \pi\tanh(a)$ (\texttt{angle\_norm}), turning the periodic encoder
into a saturating one within $(-\pi,\pi)$.

\begin{table}[t]
  \centering
  \caption{\texttt{angle\_norm} ablation (FID, 5 seeds; all six NCSN cells
  re-evaluated under one protocol). Bounding the input with $\pi\tanh$ helps
  \emph{both} cores on NCSN, so it is partly a generic input-normalization
  benefit; but the quantum gain is larger ($+4.3$ for su2 and $+8.3$ for
  RealAmplitudes vs.\ $+2.5$ for classical), consistent with an additional,
  aliasing-related failure affecting the quantum cores (Fig.~\ref{fig:gate}). On bounded DDPM the
  fix is neutral.}
  \label{tab:anglenorm}
  \small
  \begin{tabular}{llcc}
    \toprule
    Model & Core & no norm & $+\,\pi\tanh$\\
    \midrule
    NCSN & classical-SE    & $19.07\pm0.66$ & $16.62\pm0.28$\\
    NCSN & quantum-su2     & $24.54\pm0.56$ & $20.27\pm0.52$\\
    NCSN & quantum-real    & $24.52\pm0.45$ & $\mathbf{16.25\pm0.52}$\\
    DDPM & quantum-su2     & $14.52\pm0.70$ & $14.53\pm0.79$\\
    \bottomrule
  \end{tabular}
\end{table}

\paragraph{Outcome.} On NCSN, \texttt{angle\_norm} improves su2
($24.54\!\to\!20.27$) and RealAmplitudes ($24.52\!\to\!16.25$), with the
normalized RealAmplitudes ($16.25$) the lowest of all NCSN cores, below the
classical control ($16.62$). All six NCSN cells are re-evaluated under a single
protocol so the no-norm/$+\pi\tanh$ ordering is directly comparable. On bounded
DDPM the fix is neutral ($14.52\!\to\!14.53$, $\mathrm{SNR}{=}0.19$), confirming it
is safe to apply where it is not needed. Importantly, bounding the input also
\emph{improves the classical} NCSN core ($19.07\!\to\!16.62$), so $\pi\tanh$ is
partly a generic input-normalization benefit, \emph{not} a purely quantum-specific
remedy. The larger gains for the quantum cores ($+4.3$ for su2 and $+8.3$ for
RealAmplitudes vs.\ $+2.5$ for classical) are consistent with an additional
aliasing-related failure beyond the generic input-scale problem, and with
alleviating the gate collapse (Fig.~\ref{fig:gate}): the periodic encoder
suffers an aliasing failure on top of the generic input-scale issue, and the same
$\pi\tanh$ addresses both. This behavior is consistent with the $2\pi$ periodicity
of the rotation gates and is reproduced by bounding the encoder input without
otherwise changing the training configuration.

\section{Discussion}
\label{sec:disc}

\paragraph{What the no-difference result does and does not say.} At the $8$-qubit scale studied
here the quantum core is classically simulable; we therefore neither test nor
claim quantum advantage, and the comparability we observe is the expected and honest
outcome rather than a disappointment. The informative content is that, under a
strictly role-matched protocol, the variational quantum circuit is neither better
nor worse than a classical core of the same role --- it is a viable but not
superior inductive bias at this scale. This is precisely the statement that the
prevailing comparison practice, lacking matched controls, cannot make.

\paragraph{No quantum parameter-efficiency advantage is statistically established.} It is tempting to
read the $4.5$--$9\times$ fewer core parameters as efficiency, but our
parameter-matched control (Table~\ref{tab:tinymlp}) rules this out: a classical
core shrunk to the same $16$/$32$-parameter budget is numerically comparable to the quantum core,
and the $144$-parameter classical-SE is itself no better than its $32$-parameter
version. The earlier ``fewer parameters'' framing therefore conflated quantumness
with smallness against an over-provisioned control. Two further caveats apply
even to the local comparability. First, the scope is small: the core is $\sim$$10^2$
parameters out of a $\sim$23\,M-parameter model (Table~\ref{tab:params}), and on
DDPM/MNIST the SE core barely moves FID from the no-SE baseline regardless of its
form. Second, fewer parameters do not imply cheaper computation: the circuit carries
a higher per-forward-pass simulation cost than the MLP it replaces. In summary,
at this scale the variational quantum circuit reaches strict functional parity
with its parameter-matched classical counterpart while remaining \emph{neither
superior nor more efficient}; characterizing its behavior on real NISQ hardware
is left to future work.

\paragraph{Generality of the aliasing insight.} The NCSN failure and its fix are
not specific to score-based diffusion. Any pipeline that feeds an
\emph{unbounded} learned signal into a \emph{periodic} encoder --- angle
embeddings being the canonical example --- is exposed to the same wrap-around
pathology. Bounding the encoder argument (here with $\pi\tanh$) converts the
periodic map into a saturating one and removes the failure; this is a cheap,
general safeguard worth adopting whenever angle embeddings meet unnormalized
activations.

\paragraph{Limitations and outlook.} Our study is confined to $8$-qubit
simulated circuits, two small image benchmarks, and a single insertion point
(the bottleneck). Whether the picture changes with more qubits, deeper circuits,
or other insertion points is open. A natural next step is to move the trained
circuits onto real NISQ hardware and characterize the simulation-to-hardware
gap---noise sensitivity, shot noise, measurement and latency overhead, and
connectivity constraints---none of which the present classically-simulated study
captures. (At $8$ qubits the circuit remains classically simulable, so the value
of hardware runs is this sim-to-real characterization, not classical
irreproducibility.)

\section{Related Work}
\label{sec:related}

\paragraph{Quantum generative models.} A line of work explores quantum or
hybrid generative models, including quantum circuit Born
machines~\citep{liu2018qcbm,benedetti2019qcbm}, quantum
GANs~\citep{dallairedemers2018qgan,lloyd2018qgan}, and, more recently, proposals to embed
parameterized circuits into the denoiser or sampler of diffusion
models~\citep{zhang2024quddpm,parigi2024qndgdm,koelle2024qddm}. Several of these
feasibility-oriented studies report favorable numbers on small benchmarks but do
not include all of these controls simultaneously---a capacity- or role-matched
classical control in the same location, an identity-initialized scaffold, and
seed-level significance testing. Our contribution is therefore orthogonal to
proposing a new architecture: we ask how such hybrids should be \emph{evaluated},
and what a controlled evaluation actually reveals.

\paragraph{Variational circuits and feature maps.} Variational quantum
circuits~\citep{cerezo2021vqa}, angle/feature-map
embeddings~\citep{havlicek2019supervised}, circuit-centric
classifiers~\citep{schuld2020circuit}, and data
re-uploading~\citep{perezsalinas2020reuploading} form the toolbox from which our
quantum cores (RealAmplitudes, EfficientSU2) are drawn. The $2\pi$ periodicity of
the rotation gates underlying angle embedding is the property our
Section~\ref{sec:aliasing} analysis hinges on, and it connects to a broader
question of how the \emph{scale} of encoded data interacts with circuit
expressivity~\citep{perezsalinas2020reuploading}.

\paragraph{Trainability.} The trainability of VQCs is shaped by phenomena such as
barren plateaus and the choice of ansatz/entanglement~\citep{cerezo2021vqa}. Our
hardware-efficient cores~\citep{kandala2017hardware} are shallow ($\mathrm{depth}{=}2$,
$8$ qubits), so vanishing-gradient effects are mild; the failure we isolate in
Sec.~\ref{sec:aliasing} is instead an \emph{input-encoding} pathology (angle
aliasing under unbounded targets), distinct from optimization-landscape issues.

\paragraph{Fair comparison in QML.} A recurring critique of empirical QML is the
absence of strong, capacity-matched classical
baselines~\citep{schuld2022advantage,bowles2024benchmarking}, which makes reported
advantages hard to interpret. Our role-matched, identity-initialized,
multi-seed protocol is a concrete response to this critique in the generative
setting, and our SE scaffold borrows the channel-gating
idea~\citep{hu2018senet} purely as a controlled insertion point.

\section{Conclusion}
\label{sec:conc}
We presented a fair-comparison study of variational quantum circuits inside
diffusion models. Holding a squeeze-and-excitation scaffold fixed and swapping
only its core, we found no statistically detectable difference between quantum
cores and a role-matched classical control across DDPM and latent diffusion on
MNIST/CIFAR-10; the parameter-matched comparisons (same $16$/$32$-parameter
budget) show small numerical differences favoring the quantum cores, but do not
establish a statistically significant parameter-efficiency advantage at this scale. The lone
exception, score-based NCSN, revealed a concrete failure mechanism --- angle
embeddings alias when fed NCSN's unbounded score target, with input magnitudes
reaching hundreds of multiples of the $2\pi$ period --- which a simple $\pi\tanh$
normalization removes, substantially improving both quantum cores and, for
RealAmplitudes, attaining the lowest mean FID in this single-training-run
comparison (not significance-tested). We neither test nor claim quantum advantage: all circuits are
classically simulated at a small-qubit scale, and the study is designed to assess
parameterization and inductive bias rather than asymptotic speedup. The
contributions are a methodology that prevents capacity from being mistaken for
quantumness and a transferable mechanistic insight into when angle embeddings
fail. Characterizing these circuits on real
NISQ hardware is left to future work.

\section*{Acknowledgments}
This research was supported by Seoul R\&BD Program (QR250005, ``Building a
Quantum Computing--AI Integrated Development Platform and Developing Quantum--AI
Algorithm'') through the Seoul Business Agency (SBA) funded by Seoul Metropolitan
Government.

\bibliographystyle{plainnat}
\bibliography{references}

\appendix
\section{Circuit and Training Details}

\paragraph{Cores.} The \emph{RealAmplitudes} ansatz uses $\mathrm{depth}=2$ layers
of $R_Y$ rotations with circular $CZ$ entanglement ($n_q\!\cdot\!\mathrm{depth}=16$
parameters). \emph{EfficientSU2} uses $\mathrm{depth}=2$ layers of $R_Y$--$R_Z$
pairs with circular $CNOT$ entanglement ($2 n_q\!\cdot\!\mathrm{depth}=32$
parameters). Inputs are encoded by \texttt{AngleEmbedding} ($R_Y$ rotations) and
read out as $\langle Z_i\rangle$. The names \emph{RealAmplitudes} and
\emph{EfficientSU2} follow the Qiskit circuit library; we reimplement the
equivalent ans\"atze in PennyLane (\texttt{default.qubit}). The role-matched
\emph{classical} core is two blocks of $\mathrm{Linear}(8,8)\!+\!\tanh$
($2(8{\cdot}8+8)=144$ parameters). The parameter-matched (tiny) classical
controls use a bias-free low-rank map $f(x)=W_2\tanh(W_1 x)$ with
$W_1\in\mathbb{R}^{r\times 8}$, $W_2\in\mathbb{R}^{8\times r}$, giving $16r$
parameters ($16$ for rank $r{=}1$, $32$ for $r{=}2$), matching the $16$/$32$
parameter counts of RealAmplitudes/EfficientSU2.

\paragraph{Hyperparameters.} Optimizer Adam ($10^{-4}$); batch $64$; $300$ epochs;
cosine schedule; gradient clip $1.0$; EMA decay $0.9999$. Quantum simulation:
PennyLane \texttt{default.qubit}, $\mathrm{diff\_method}=\texttt{backprop}$.

\paragraph{Architectures and schedules.} The backbone is a U-Net with GroupNorm,
SiLU, and sinusoidal timestep embedding; the SE core is inserted at the two
bottleneck (\texttt{middle}) blocks, where channel statistics are pooled to a
length-$8$ vector before encoding. DDPM uses a cosine $\beta$-schedule with
$T{=}1000$ and ancestral sampling; NCSN uses a geometric noise scale schedule
($\sigma_{\min}{=}0.01$, $\sigma_{\max}{=}25$, $200$ scales) trained by denoising
score matching and sampled with annealed Langevin dynamics ($10$ steps per scale,
step size $\epsilon{=}2\!\times\!10^{-5}$, a final denoising step); LDM runs DDPM in the latent space of a
pretrained autoencoder ($\mathrm{latent\_dim}{=}4$). The LDM/CIFAR-10 setting is
included as an architecture-family diversity check, not a state-of-the-art
latent-diffusion baseline; its relatively high FID ($\sim$99) is limited by this
compact low-dimensional autoencoder---not by the SE core under study, which is
the controlled variable and where quantum/classical remain comparable. Images are normalized to
the network's input range; FID/SSIM are computed on samples denormalized to
$[0,1]$. The quantum core is evaluated as a single batched circuit per
bottleneck block on CPU; its expectation values are computed analytically
(\texttt{shots=None}) from the statevector (PennyLane \texttt{default.qubit}),
not sampled, so there is no shot noise. All other computation runs on GPU.

\begin{table}[h]
  \centering
  \caption{Full SSIM (mean$\pm$std, 5 seeds; higher is better). Differences
  between cores are within noise, consistent with the comparable FID results in
  Table~\ref{tab:main}.}
  \label{tab:ssim}
  \small
  \begin{tabular}{llcccc}
    \toprule
    Model & Dataset & none & classical-SE & quantum-real & quantum-su2\\
    \midrule
    DDPM & MNIST   & $0.3707\pm0.0023$ & $0.3703\pm0.0024$ & $0.3707\pm0.0019$ & $0.3719\pm0.0019$\\
    DDPM & CIFAR-10& $0.2307\pm0.0015$ & $0.2311\pm0.0013$ & $0.2311\pm0.0016$ & $0.2355\pm0.0015$\\
    LDM  & MNIST   & $0.3749\pm0.0020$ & $0.3772\pm0.0014$ & $0.3761\pm0.0011$ & $0.3761\pm0.0022$\\
    LDM  & CIFAR-10& $0.2397\pm0.0025$ & $0.2406\pm0.0020$ & $0.2399\pm0.0013$ & $0.2389\pm0.0007$\\
    NCSN & MNIST   & $0.3907\pm0.0017$ & $0.3819\pm0.0024$ & $0.3778\pm0.0023$ & $0.3799\pm0.0025$\\
    \bottomrule
  \end{tabular}
\end{table}

\end{document}